\documentclass[letterpaper,10pt,conference]{ieeeconf}
\IEEEoverridecommandlockouts       
\usepackage{amssymb}
\usepackage{amsmath}
\usepackage[english]{babel}
\usepackage{float}
\usepackage{subcaption}
\usepackage{graphicx}
\usepackage{hyperref}
\usepackage{mathtools}

\usepackage{filecontents}
\usepackage[noadjust]{cite}
\usepackage[font=footnotesize,labelfont=footnotesize]{subcaption}
\usepackage[font=footnotesize,labelfont=footnotesize]{caption}
\usepackage{comment}
\usepackage{authblk}
\usepackage{multirow}
\usepackage{xcolor}
\usepackage{algorithm}
\usepackage{algorithmic}

\DeclareMathOperator*{\argmin}{arg\,min}

\definecolor{darkblue}{rgb}{0.15,0.15,0.55}
\definecolor{lightgrey}{rgb}{0.75,0.75,0.75}

\providecommand{\codecomment}[1]{\textcolor{lightgrey}{\dotfill//\,}\textcolor{darkblue}{\textrm{#1}}}
\newcommand{\qedm}{\hfill $\square$}

\begin{document}
\title{\LARGE \bf Efficient Obstacle Rearrangement for Object Manipulation Tasks in Cluttered Environments
}

\author{Jinhwi Lee$^{1,2}$, Younggil Cho$^{1}$, Changjoo Nam$^{1}$, Jonghyeon Park$^{3}$ and Changhwan Kim$^{1,*}$

\thanks{This work was supported by the Technology Innovation Program and Industrial Strategic Technology Development Program (10077538, Development of manipulation technologies in social contexts for human-care service robots).
$^{1}$Korea Institute of Science and Technology, Seoul, Korea. Emails: {\tt\small \{jinhooi, briancho, cjnam, ckim\}@kist.re.kr}. $^{2}$Graduate School of Hanyang University, Seoul, Korea. $^{3}$Division of Mechanical Engineering, Hanyang University, Seoul, Korea. {\tt\small jongpark@hanyang.ac.kr}. *Corresponding author.}
}


\maketitle

\begin{abstract}
We present an algorithm that produces a plan for relocating obstacles in order to grasp a target in clutter by a robotic manipulator without collisions. We consider configurations where objects are densely populated in a constrained and confined space. Thus, there exists no collision-free path for the manipulator without relocating obstacles. Since the problem of planning for object rearrangement has shown to be NP-hard, it is difficult to perform manipulation tasks efficiently which could frequently happen in service domains (e.g., taking out a target from a shelf or a fridge).  

Our proposed planner employs a collision avoidance scheme which has been widely used in mobile robot navigation. The planner determines an obstacle to be removed quickly in real time. It also can deal with dynamic changes in the configuration (e.g., changes in object poses). Our method is shown to be complete and runs in polynomial time. Experimental results in a realistic simulated environment show that our method improves up to 31\% of the execution time compared to other competitors.

\end{abstract}

\section{Introduction}
Robotic manipulation has long been studied and applied in the context of industrial applications. As a result, a significant portion of manufacturing processes can be automated by the advances in robot technologies such as mechanism design and control. However, object manipulation in service domains has not succeeded sufficiently owing to the unstructured environments which could incur uncertainties and contingencies. Industrial manipulators are not appropriate for operations in such environments as they oftentimes assume an environment that is structured, thus certain and static.

For motion planning problems of manipulators in unstructured environments, it is important to consider task-level planning because motion planning of the end-effector for itself may not be able to deal with complex situations like grasping a target object from clutter where many obstacles occlude the target. Thus, task and motion planning (TAMP) has addressed the problems where high-level task planning and low-level motion planning need to be orchestrated~\cite{kaelbling2013integrated,lozano2014constraint,srivastava2014combined}. For example, grasping a target from clutter needs multiple subtasks in conjunction with motion planning where subtasks relocate obstacles which prevent the end-effector from reaching to the target. However, planning for such the subtasks could be computationally intractable since rearrangement of obstacles is shown to be NP-hard~\cite{wilfong1991motion,stilman2008planning}.

\begin{figure}[t!]
\captionsetup{skip=0pt}
    \centering
   	\includegraphics[scale=0.11]{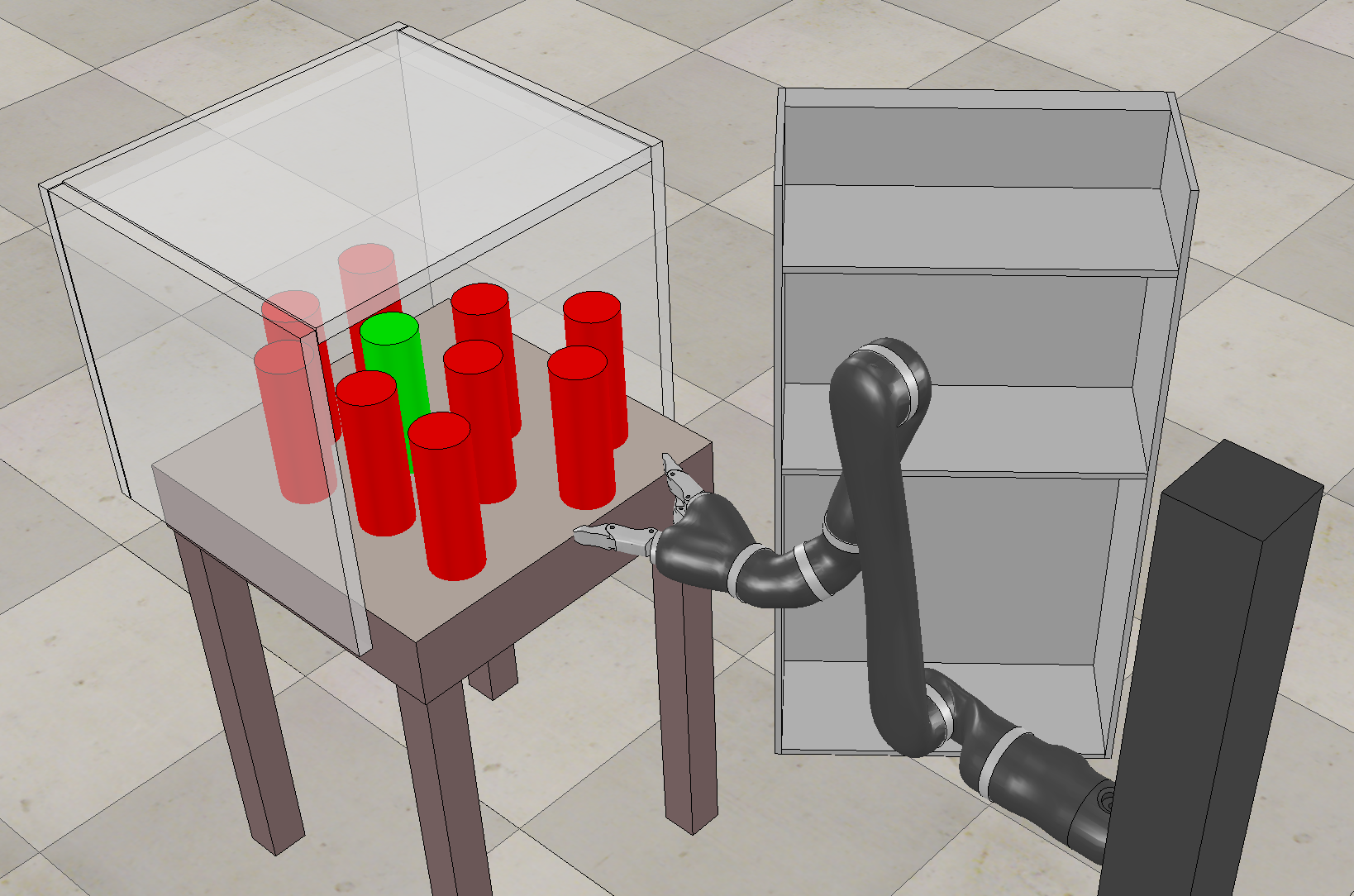}
    \caption{An example of the grasping task in clutter. The target object (green) is surrounded by obstacles (red) on the table which is constrained by walls and the ceiling. Relocating obstacles is necessary to reach to the target.}
    \label{fig:problem}
    \vspace{-23pt}
\end{figure}

In this paper, we study the task and motion planning problem where robots need to rearrange obstacles in order to accomplish manipulation tasks which could frequently happen in service domains. We consider complex and cluttered environments in everyday life such as a fridge or a shelf populated with various objects arranged irregularly. Thus, a target object is not accessible to the manipulator without removing some obstacles as shown in Fig.~\ref{fig:problem}. In such environments, simple motion planning generating paths to the target object would not be able to solve the problem, but the robots need to plan for both task and motion to determine what to relocate until the target becomes accessible. 

We present a complete and polynomial-time algorithm that decides obstacles to be removed to grasp the target without collisions. An important aspect of obstacle rearrangement is determining the direction to approach to the target because the direction decides what to remove. However, finding the direction is a difficult problem as one needs to consider constraints like robot kinematics as well as efficiency of rearrangement (e.g., minimizing the number of obstacles to be removed). We tackle this problem by employing a collision avoidance scheme widely used in mobile robot navigation. Although the scheme has been proposed for low-level motion planning, our modification and use of it enables decision-making of manipulators for task planning in cluttered environments. 

The following are the contribution of this work:

\begin{itemize}
    \item We propose a fast real-time algorithm determining obstacles to be relocated. The algorithm is parameter-free so users can reduce the laborious manual search to find parameter values. 
    \item The algorithm is shown to be complete and runs in polynomial-time. We provide proofs for completeness and polynomial running time.
    \item We provide experimental results in a realistic simulated environment which show that our method outperforms other existing competitors.
    \item The method can deal with dynamic changes since replanning for an updated configuration can be done quickly.
\end{itemize}

\section{Related Work}

In order to find an obstacle to be removed, Zacharias et al.~\cite{zacharias2006bridging} use a variant of Vector Field Histogram (VFH) which uses Gaussian distributions. A distribution represents the density of obstacles around a target. The obstacles in the directions with low densities in the distribution are removed sequentially since the directions with low densities mean sparsely populated areas so fewer obstacles need to be removed. The distribution is updated periodically to reflect objects that are removed. If any density of the distribution goes below a predetermined threshold value, it indicates that the target object is no longer blocked by obstacles in that direction. Thus, the manipulator can grasp the target without collisions. Our algorithm is inspired by this work but it requires adjustments of the threshold value if the configuration of objects changes. However, adjusting the threshold value is not done by a principled method but needs to be done empirically. Therefore, this method has limited applicability in varied environments and configurations which are common in service domains.

In~\cite{dogar2012planning}, a planning framework that utilizes non-prehensile actions to grasp an object in clutter is proposed. The framework does not aim to minimize the number of obstacles to be removed which often influences the execution time of grasping tasks. It removes obstacles that the end-effector could collide with if the end-effector follows the straight path between the target and the initial pose. Although this method generates the distance-optimal path of the end-effector, some obstacles could have to be removed unnecessarily. In addition, the framework does not consider dynamic environments so grasping would fail if the poses of objects change during execution.

Several recent work considering object manipulation in clutter~\cite{moll2018randomized,srivastava2014combined,haustein2015kinodynamic} also do not directly optimize energy or time used for accomplishing grasping tasks but mainly concern about validity of their plans. For example, \cite{moll2018randomized} presents a randomized motion planner to grasp an object in clutter where no collision-free path exists in the initial configuration. The planner generates a path that the manipulator follows while pushing obstacles along the path. However, the path is generated from a random tree expansion whose strategy only concerns about improving the coverage of the tree rather than the number of obstacles to be pushed.


Among these work, none of them has directly tackled the problem of minimizing the number of obstacles to be relocated in dynamic environments. By considering a more appropriate and practical objective value, we aim to perform the grasping task more efficiently in reduced execution time.

\section{Problem Description}
We consider the problem of grasping a target object that requires rearrangement of obstacles to avoid collisions. We assume that the robot knows the configuration of objects where there could be dynamic changes in the object locations. A configuration has $N$ objects including target in a planar 2D space. The objects are densely populated in the space so there is no collision-free path to the target object.

In this setting, we aim to grasp the target object while considering costs for execution where the cost can be defined as time (or energy consumption equivalently). Obstacle relocation often dominates the overall execution time since grasping and releasing obstacles takes longer than transporting them. Thus, we aim to minimize the number of obstacles to be relocated to reduce the execution time for grasping. More formally, we find $o_i \in O$ for $i=1, 2,\cdots, k$ where $o_i$ is the $i$-th object to be relocated, $O$ is the set of objects in a reachable workspace of the manipulator, and $k$ is the number of object to be relocated. Therefore, minimizing $k$ is the objective of the problem.

\section{A Method for Efficient Obstacle Rearrangement}
\label{sec:approach}

\begin{figure}[t!]
\captionsetup{skip=0pt}
    \centering
    \begin{subfigure}[b]{0.21\textwidth}
    \captionsetup{skip=0pt}
   	\includegraphics[width=\textwidth]{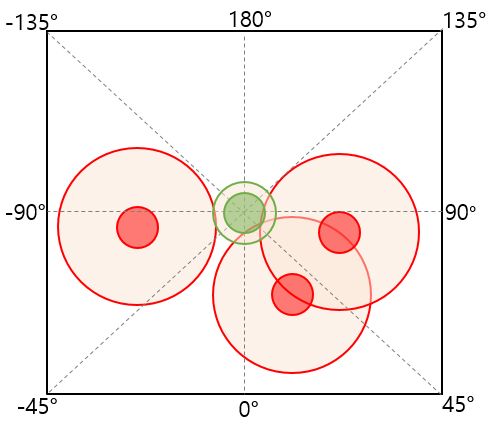}
    \caption{An example configuration} \label{fig:polara}
    \end{subfigure}
    
    \begin{subfigure}[b]{0.2\textwidth}
    \captionsetup{skip=0pt}
   	\includegraphics[width=\textwidth]{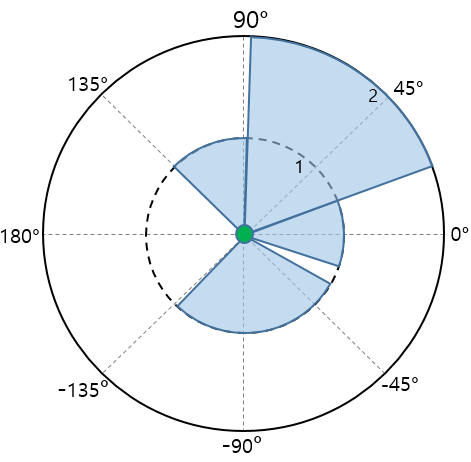}
    \caption{The polar histogram of the configuration shown in (a)} \label{fig:polarb}
    \end{subfigure}
    \begin{subfigure}[b]{0.24\textwidth}
    \captionsetup{skip=0pt}
   	\includegraphics[width=\textwidth]{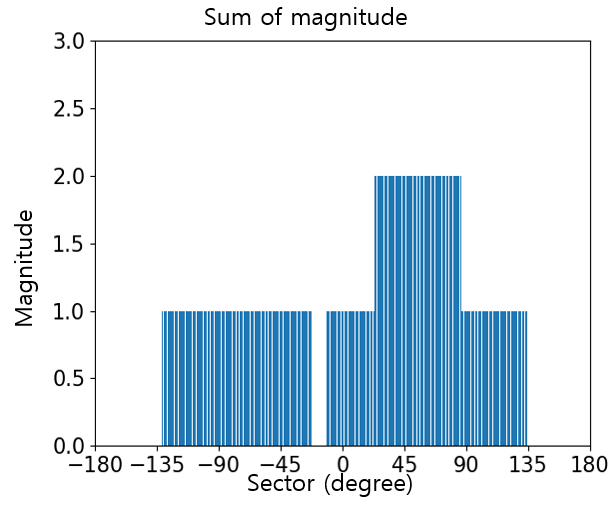}
    \caption{The normal histogram which is equivalent to the polar histogram (b)} \label{fig:polarc}
    \end{subfigure}  
    \caption{An example configuration (green: robot, red: obstacle) and its corresponding histograms with different representations.} \label{fig:polarhistogram}\vspace{-20pt}
\end{figure}

The direction of the end-effector approaching to the target plays an important role in grasping objects in clutter since accessibility of the target could change depending on the approaching direction. In mobile robot navigation, Vector Field Histogram+ (VFH+)~\cite{ulrich1998vfh+} has been used widely to determine the driving direction of robots to avoid collisions with obstacles. The original VFH computes a polar histogram according to the positions and distances of obstacles around a mobile robot as shown in Fig.~\ref{fig:polarhistogram}. Then, the robot determines the direction to proceed by choosing a sector (see the $x$-axis in Fig.~\ref{fig:polarc}) whose vector magnitude of the histogram ($y$ values in Fig.~\ref{fig:polarc}) is smaller than a threshold value. An improved version, VFH+ adds another threshold value to make the histogram simpler so a faster determination of the driving direction is allowed.


Our idea is to use VFH+, which is a local path planner, for subtask planning of manipulators which is determining obstacles to be rearranged. However, applying the method directly to our problem would incur few problems as it has been developed for mobile robot navigation. First, VFH+ simplifies the original polar histogram (i.e., regards possibly multiple obstacles clustered as a single large obstacle)
so cannot distinguish various objects. This is not a serious problem in mobile robot navigation since a mobile robot just need to recognize the direction blocked by obstacles. However, we need to distinguish different obstacles since they need to be relocated separately because the end-effector cannot grasp them at once. Second, VFH+ could find a driving direction which would produce a long detour globally, which a limitation of local planners. To deal with the two problems, we modify VFH+ as described in this section.

\subsection{Determining What to Remove using Modified VFH+}
\label{sec:vfh}

We describe a modified version of VFH+ in order for the robot to recognize the environment to choose the obstacles to be relocated. The key idea of our modification is (1) eliminating threshold values required in VFH+ and (2) considering the environment in the perspective of the target but not that of the robot. While the canonical use of VFH+ is that a mobile robot uses the histogram to find obstacle-free directions, we use it in an inverse fashion to find obstacle-free directions from the target object. 

Since VFH+ is a local path planner, it generate paths within a local map. Given a square map\footnote{This is without loss of generality as a map with other shapes (e.g., a rectangle, any irregular shape) can be represented by a square that circumscribes the shape.} whose size is $w_s$, VFH+ needs two constants $a, b \in \mathbb{R}^+$ to determine the vector magnitude, which represents density of obstacles in the histogram. They are determined by the following relationship
\begin{equation} \label{findab}
a-b\left(\frac{w_s-1}{2}\right)^2=1.
\end{equation}
The constants do not need to be fine-tuned. The only condition to be satisfied is that $a,  b > 0$.

\begin{figure}[t!]
\captionsetup{skip=0pt}
    \centering
   	\includegraphics[scale=0.36]{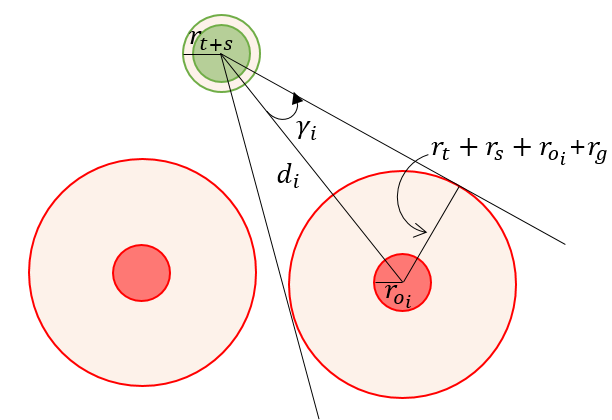}
    \caption{The enlargement process of the target (green) and obstacles (red) to generate a configuration space. The target with radius $r_t$ is expanded by $r_s$ which is the safety margin (the larger circle on the target). Each obstacle with its radius $r_o$ is expanded by the target radius $r_t$ with the safety margin $r_s$ and the grasper radius $r_g$.}
    \label{fig:VFH1}\vspace{-20pt}
\end{figure}

On the other hand, the size of the end-effector and the objects should be taken into account in the histogram computation to prevent collisions while performing grasping tasks. Thus, we generate a configuration space (C-space) considering the volume/size of objects and the end-effector as follows. First, the target with radius $r_t$ is expanded by $r_s$ which is the safety margin. The safety margin is needed to let the end-effector put a little space between obstacles and the end-effector itself grasping the target. Thus, any possible contact can be avoided while the target is being transported. Each obstacle with its own radius $r_o$ is expanded by the target radius $r_t$ with the safety margin $r_s$ and the end-effector radius $r_g$. Fig.~\ref{fig:VFH1} describes how the expansion is applied to the target (green) and an obstacle (red). Therefore, an obstacle in a C-space has a radius $r_{total} = r_t + r_s + r_o + r_g$. Notice that our modeling of objects as circles can be generalized to other asymmetric shapes by representing them with circumcircles.

For each obstacle, the enlargement angle $\gamma$ is defined as 
\begin{equation} \label{checkvolume}
\gamma_{i}=\frac{1}{\alpha}\sin^{-1}\frac{r_{total}}{d_{i}}
\end{equation}
where $i$ is the index of the obstacle $o_i$, $d_{i}$ is the distance from the target to $o_i$, and $\alpha$ is the angular resolution which represents the degree to divide the sectors of the polar histogram (the smaller the finer representation of obstacles). In this work, $\alpha = 1$, which means that a vector magnitude is computed for each degree so the obstacles are represented very precisely in the histogram. 

For each sector in the histogram, the polar histogram $H_{i}$ of $o_i$ is calculated by
\begin{equation} \label{histogram}
H_{i}=\begin{cases}
(c_{i})^2(a-bd_{i}^2), & \mbox{if }z\in [\beta_i-\frac{\gamma_i}{\alpha}, \beta_i+\frac{\gamma_i}{\alpha}]\\
0 & \mbox{otherwise} 
\end{cases}
\end{equation}
where $c_{i}$ is the coefficient determining the certainty of sensor readings, $\beta_i$ is angle of target to $o_i$ and $z$ is the angular sector of $o_i$ (i.e., the direction that $o_i$ is located at) in the histogram. In this work, we consider certain environments so $c_i = 1$.

Finally, the overall histogram considering all obstacles is computed by the sum of the polar histogram of each obstacle that is
\begin{equation} \label{sumhistogram}
H=\sum_{i}H_{i}.
\end{equation}

The histogram represents how the objects are configured around the target object.
We note that the difference from the basic VFH+ to our modification is that our method does not simplify the histogram so can distinguish different obstacles minutely to grasp them separately. In addition, our method does not use any threshold values which are required in the original VFH+.

\subsection{The Algorithms}

Alg.~\ref{alg:task} describes the task planning process for removing obstacles occluding the target. The algorithm receives the following information as input: (1) target object $t$, (2) the configuration $O$ of all objects in the workspace (i.e., the reachable area of the end-effector), and robot configuration information $M$ (i.e., the kinematics and the base pose). We assume that the information regarding the objects such as the radii and centroids can be perceived from a vision system. The algorithm returns the object to be grasped $o_r$, which is either the target or the obstacle blocking the target.

\begin{figure}[t]
\captionsetup{skip=0pt}
    \centering
    \begin{subfigure}[t]{0.22\textwidth}
    \captionsetup{skip=0pt}
   	\includegraphics[width=\textwidth]{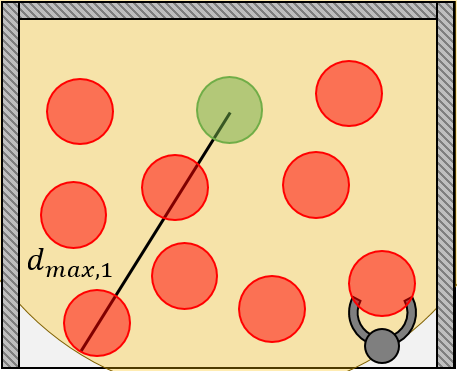}
    \caption{The histogram is calculated for the space whose radius is $d_{\mbox{\scriptsize max}}$.} \label{fig:alg1a}
    \end{subfigure}
    \hspace*{\fill} 
    \begin{subfigure}[t]{0.22\textwidth}
    \captionsetup{skip=0pt}
   	\includegraphics[width=\textwidth]{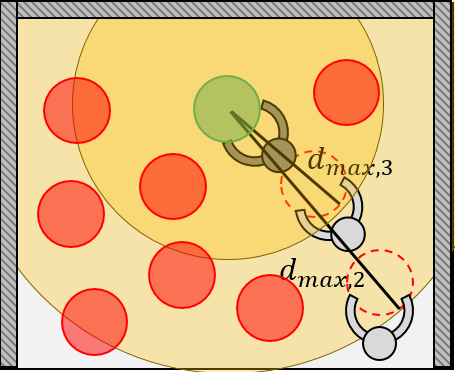}
    \caption{The histogram is calculated within a newly updated $d_{\mbox{\scriptsize max},i}$ for the current configuration.} \label{fig:alg1b}
    \end{subfigure}    
    \caption{An illustration of our method. In Alg.~\ref{alg:task}, $d_{\mbox{\scriptsize max}}$ is updated as the configuration changes by a sequence of object removals. In (a), the object to be relocated is determined by the direction chosen using the histogram in Alg.~\ref{alg:acc}. In (b), Alg.~\ref{alg:acc} runs with an updated $d_{\mbox{\scriptsize max}}$ which is reduced to the previously removed object location.
    } \label{fig:alg1}\vspace{-10pt}
\end{figure}

The maximum Euclidean distance $d_{\mbox{\scriptsize max}}$ among all distances calculated between the target object and each of the other objects is chosen in the current configuration (lines 3--8). 
The distance $d_{\mbox{\scriptsize max}}$ is updated as the configuration changes by a sequence of object removals as shown in Fig.~\ref{fig:alg1}. 
If $d_{\mbox{\scriptsize max}}$ does not change, it means that there exists a path taking a detour around obstacles surrounding the target. However, the histogram may not show a sector with zero magnitude because the histogram does not take into account such a path which is not straight between the target and the end-effector as described in Fig.~\ref{fig:polarb}.
Based on this information and other input arguments, the algorithm finds the object to be removed by invoking Alg.~\ref{alg:acc} (line 9). If the object to be removed $o_r$ chosen by Alg.~\ref{alg:acc} is the target (i.e., the target is directly accessible without removing any obstacle), the object is grasped and removed from $O$ (lines 11--12). Then the algorithm terminates as $t$ is not in $O$ anymore. If Alg.~\ref{alg:acc} returns an object $o_r$ that is not the target, it means that the target is not accessible directly and $o_r$ blocks the target. Thus, Alg.~\ref{alg:task} runs recursively until all obstacles blocking the target are removed so the target is accessible to the end-effector (line 14). Finally, the algorithm returns the configuration $O$ where $t$ is removed.

Alg.~\ref{alg:acc} illustrates the accessibility check done using the modified VFH+.  Alg.~\ref{alg:acc} returns the object $o_r$ to be removed. It computes the polar histogram using the modified VFH+ (lines 1--8). Then the object that is with the lowest vector magnitude in the histogram (i.e., the most accessible around the target) is selected (lines 9--13).

\begin{figure}[t]
\captionsetup{skip=0pt}
    \centering
    \begin{subfigure}[b]{0.49\textwidth}
    \captionsetup{skip=0pt}
   	\includegraphics[width=\textwidth]{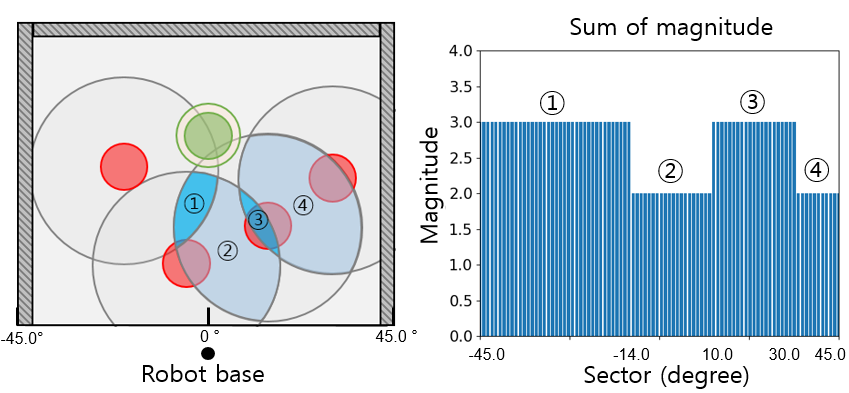}
    \caption{The case where no obstacle-free direction. The histogram on the right does not have sectors with zero vector magnitude. Thus the manipulator cannot reach to the target without removing obstacles.} \label{fig:VFH2a}
    \end{subfigure}
    \begin{subfigure}[b]{0.49\textwidth}
    \captionsetup{skip=0pt}
   	\includegraphics[width=\textwidth]{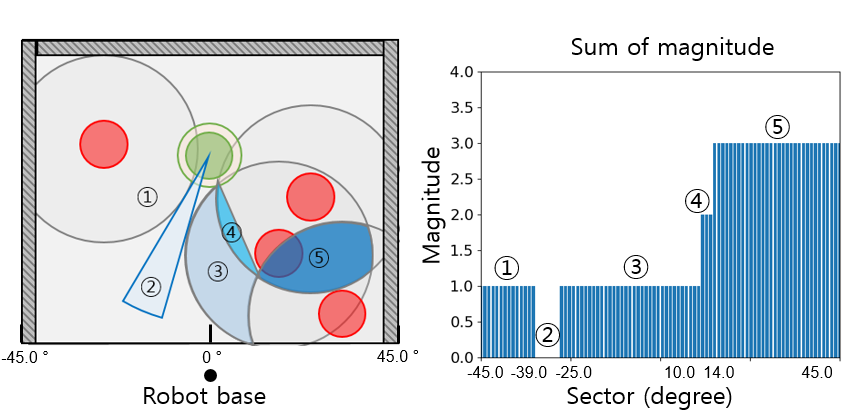}
    \caption{The case where there is an obstacle-free direction. The histogram on the right shows that Area 2 has zero vector magnitude. Therefore, the manipulator reach to the target directly through Area 2.} \label{fig:VFH2b}
    \end{subfigure}    
    \caption{An example illustrating the histograms (right) computed by the modified VFH+ for the cluttered configurations in the left. The numbers in circles in the left represent areas in the configurations. The numbers in the histograms matches to the areas in the right. From the target object (green), the downward direction to the robot base (the position of the shoulder of the manipulator) is $0^{\circ}$, minus and plus values indicate the left and the right side, respectively. In this case, the histograms are shown from $-45^{\circ}$ to $45^{\circ}$ because outside of the range is out of the workspace of the robot.} \label{fig:VFH2}\vspace{-10pt}
\end{figure}

\renewcommand{\algorithmicrequire}{\textbf{Input:}}
\renewcommand{\algorithmicensure}{\textbf{Output:}}
\renewcommand{\algorithmiccomment}[1]{\bgroup\hfill//~#1\egroup}
\begin{algorithm}
\caption{{\scshape TaskPlanner}} \label{alg:task}
\begin{algorithmic}[1]
{\small
\REQUIRE Target $t$, configuration $O$ of all objects in the workspace including $t$, robot configuration information $M$
\ENSURE The configuration $O$ after grasping $t$

\STATE $d_{\mbox{\scriptsize max}}=0$
\WHILE{$t \in O$} 
    \FOR{each $o_i \in O$} 
        \STATE $d=$ {\scshape EuclideanDist}($o_i,t$)
        \IF{$d>d_{\mbox{\scriptsize max}}$}
            \STATE $d_{\mbox{\scriptsize max}}=d$
        \ENDIF
    \ENDFOR
    \STATE $o_r=$ {\scshape AccessibilityCheck}($O,t,M,d_{\mbox{\scriptsize max}}$) \codecomment{Determine the object to be removed}
    \IF{$o_r==t$} 
        \STATE Grasp $o_r$ \codecomment{Grasp the target}
        \STATE $O=O \setminus o_r$
    \ELSE 
        \STATE $O$ = {\scshape TaskPlanner}$(o_r, O, M)$ \codecomment{Remove the obstacle occluding the target recursively until all obstacles are removed}
    \ENDIF
\ENDWHILE
\RETURN $O$}
\end{algorithmic}
\end{algorithm}

\begin{algorithm} 
\caption{{\scshape AccessibilityCheck}} \label{alg:acc}
\begin{algorithmic}[1] 
{\small
\REQUIRE Configuration $O$ of all objects in the workspace, target $t$, workspace distance $d_{\mbox{\scriptsize max}}$, robot configuration information $M$
\ENSURE The object $o_r$ to be removed 

\STATE $R=$ {\scshape RotationMatrix}$(t,M)$ \codecomment{Calculate rotation matrix from $t$ to base position of $M$}
\STATE $O = O \setminus t$
\STATE $O^{\prime}=$ {\scshape FindObj}($O,t,d_{\mbox{\scriptsize max}}$) \codecomment{Find all objects in $O$ whose distance from the target is smaller than or equal to $d_{\mbox{\scriptsize max}}$}
\FOR{$o_i \in O^{\prime}$}
    \STATE $o_i=R*o_i$ \codecomment{Rotate $o_i$ based on $0^\circ$ from target to the robot base}
    \STATE $\beta _i=$ {\scshape Angle}$(t,o_i)$ \codecomment{Calculate angle from $t$ to $o_i$}
\ENDFOR
\STATE $H=$  {\scshape ModifiedVFH+}$(O^{\prime},t)$ \codecomment{Modified VFH+ described in Sec.~\ref{sec:vfh}}
\FOR{$o_i \in O^{\prime}$}
    \STATE $C_i=\left\vert \beta_i-\argmin_{\theta} H \right\vert$ \codecomment{Compute $C_i$ which is degree between $beta_i$ and $\argmin_{\theta} H$}
\ENDFOR
\STATE $o_r=\argmin_{o \in O^\prime}(C_i)$ \codecomment{Select $o_i$ based on minimum $C_i$}
\RETURN $o_r$
}
\end{algorithmic}
\end{algorithm}

We note that Alg.~\ref{alg:task} is complete and runs in polynomial time as shown in the following theorems.
\smallskip

\noindent \textbf{Lemma 1.} Alg.~\ref{alg:acc} is complete so always returns an object to be grasped.

\noindent \textbf{Proof.} The modified VFH+ computes the sum of histograms of each obstacle in the current configuration. The overall histogram must have at least one minimum value of the vector magnitude (e.g., 2 and 4 in Fig.~\ref{fig:VFH2a} and 2 in Fig.~\ref{fig:VFH2b}). There are two cases depending on the minimum value of the vector magnitude: (i) If the value is zero, there is a direction to the target which is obstacle-free. Thus, the algorithm can return the target which is to be grasped. (ii) If the value is nonzero, there exists at least one obstacle blocking the target. Thus, the algorithm returns the obstacle to be removed. In any case, the algorithm returns an object to be grasped. \qedm

\noindent \textbf{Theorem 2.} Alg.~\ref{alg:task} is complete so eventually grasps the target.

\noindent \textbf{Proof.} By Lemma 1, Alg.~\ref{alg:acc} is complete so always returns an object to be grasped (either the target or an obstacle). Therefore, Alg.~\ref{alg:task} always can grasp the object returned by  Alg.~\ref{alg:acc}. In each recursive call of Alg.~\ref{alg:task}, one object is removed. Once the recursion repeats for all $N-1$ objects in the worst case (i.e., all obstacles are blocking the target so removed), the last remaining object, which is the target, is grasped. Thus, Alg.~\ref{alg:task} is guaranteed to grasp the target.
\qedm
\smallskip

\noindent \textbf{Theorem 3.} Alg.~\ref{alg:task} runs in polynomial time. 

\noindent \textbf{Proof.} Alg.~\ref{alg:task} is called recursively at most $N$ times for all $N$ objects in $O$. Computing $d_{\mbox{\scriptsize max}}$ takes $O(N)$. The subroutine Alg.~\ref{alg:acc} is with $O(N)$ because \textsc{ModifiedVFH+} has $O(N + (N-1))$ to run two loops with at most $N$ iterations and to compute histograms of all $N-1$ obstacles. Therefore, the time complexity of Alg.~\ref{alg:task} is $O(N(N+N)) =  O(N^2)$.
\qedm
\smallskip

\section{Experiments}
\label{sec:exp}

We run experiments in a realistic simulated environment using a high fidelity robotic simulator V-REP~\cite{rohmer2013v} where dynamics can be modeled by different physics engines (we used Vortex Dynamics). Kinova JACO1, a 6-DOF manipulator anchored at a base location, is used in the experiments. We use BIT-RRT~\cite{devaurs2013enhancing} implemented in Open Motion Planning Library (OMPL)~\cite{sucan2012open} for motion planning of the end-effector. The test instances of the problem are with cluttered and constrained environments like shelves or fridges. Thus, we assume that the manipulator only can approach to the objects from one side as shown in Fig.~\ref{fig:problem}. The grasped obstacles are placed in any available empty space nearby the manipulator.

We test instances with $N = 2, 4, 6, 8, 10$ where $N$ includes the target object. The objects are placed in a 0.5\,m by 0.5\,m planar space randomly. The diameters of the objects are also randomly sampled from $\mathcal{U}(6, 7.5)$ whose unit is centimeter. The random generation of problem instances is designed to produce high-density configurations of objects in a small confined space.

We compare our method to other competitive methods. The baseline method is developed in~\cite{dogar2012planning} in which the end-effector removes the obstacles that are on the distance optimal (i.e., straight) path from the base pose of the end-effector to the target. Since it does not aim to minimize the number of objects to be removed to find a collision-free path, the time for completing the grasping task would take unnecessarily long time. Another comparison is done with the method presented in~\cite{zacharias2006bridging} which uses Gaussian distributions to find the obstacle to be removed. The direction (angle) deciding the obstacle to be removed is selected using a threshold value applied to the Gaussian distribution. The threshold value should be determined empirically, so there is no principled way for choosing the value (it is hand-tuned in the experiment to produce the best performance).

\begin{figure}[h!]
\captionsetup{skip=0pt}
    \centering
    \begin{subfigure}[b]{0.35\textwidth}
    \captionsetup{skip=0pt}
   	\includegraphics[width=\textwidth]{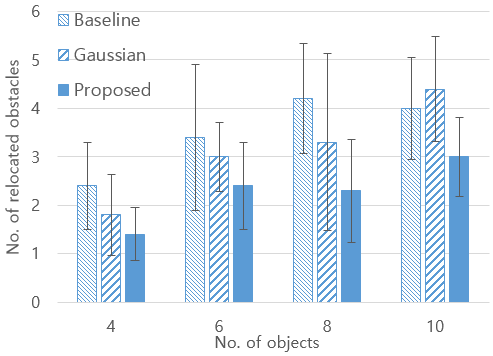}
    \caption{The averages of the number of objects relocated and standard deviation from $N=4$ to $N=10$.} \label{fig:obs}
    \end{subfigure}
    \vskip\baselineskip
    \begin{subfigure}[b]{0.35\textwidth}
    \captionsetup{skip=0pt}
   	\includegraphics[width=\textwidth]{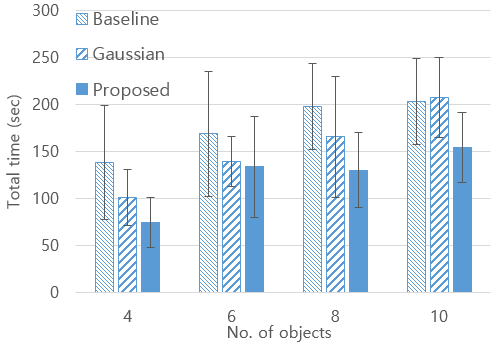}
    \caption{The averages of execution time and standard deviation from $N=4$ to $N=10$} \label{fig:time}
    \end{subfigure}
    \caption{The result of verifying the our algorithm while changing the number of objects $N$ from 4 to 10 including the target object. (a) represents the number of objects relocated by $N$. (b) represents the execution time by $N$} \label{fig:data} \vspace{-10pt}
\end{figure}

We measure the execution time of completing a grasping task (from the time the end-effector starts from its base to the time the target is taken out to the base). Note that planning time for determining one object is about 0.1 sec in  a system with Intel Core i7 4.20GHz and 8GB RAM which is suitable for real-time applications. We also measure the number of obstacles that are removed in total. We generate 10 different random configurations and run the three methods in each of the random instances (i.e., they are compared in the same configuration). The results, average values and standard deviations computed from the repetitions, are shown in Fig.~\ref{fig:data} and Table~\ref{tab:3}.

\begin{table}[h!]
\centering
\begin{subtable}[t]{0.95\linewidth}
\captionsetup{skip=1pt}
\centering
\scalebox{0.95}{%
\begin{tabular}{|c|c|c|c|c|}
\hline
\multirow{2}{*}{\begin{tabular}[c]{@{}c@{}}Method\end{tabular}} & \multicolumn{4}{c|}{No. of objects}                           \\ \cline{2-5} 
                                                                             & 4           & 6           & 8           & 10          \\ \hline
Baseline                                                                     & 138.2 (60.6) & 168.6 (67.0) & 198.0 (45.7) & 203.4 (45.7) \\ \hline
Gaussian                                                                     & 100.9 (29.7) & 139.2 (27.0) & 165.6 (64.1) & 207.6 (42.8) \\ \hline
Proposed                                                                     & 74.6 (26.6)  & 133.8 (53.5) & 130.2 (39.5) & 154.8 (37.3) \\ \hline
\end{tabular}%
}
\caption{The average execution time (sec) and standard deviation} 
\label{tab:1}
\end{subtable}\vspace{6pt}

\begin{subtable}[t]{0.95\linewidth}
\captionsetup{skip=1pt}
\centering
\scalebox{0.95}{%
\begin{tabular}{|c|c|c|c|c|}
\hline
\multirow{2}{*}{\begin{tabular}[c]{@{}c@{}}Method\end{tabular}} & \multicolumn{4}{c|}{No. of objects}            \\ \cline{2-5} 
                                                                                       & 4       & 6        & 8      & 10      \\ \hline
Baseline                                                                               & 2.4 (0.9) & 3.4 (1.5)  & 3.8 (1.1) & 4.0 (1.1) \\ \hline
Gaussian                                                                               & 1.8 (0.8) & 3.0 (0.7) & 3.3 (1.8) & 4.4 (1.1) \\ \hline
Proposed                                                                               & 1.4 (0.6) & 2.4 (0.9)  & 2.3 (1.1) & 3.0 (0.8) \\ \hline
\end{tabular}%
}
\caption{The average number of relocated obstacles and standard deviation} 
\label{tab:2}
\end{subtable}
\caption{The results of experiments over 10 repetitions}
\label{tab:3} \vspace{-25pt}
\end{table}

\begin{figure*}[t!]
    \centering
   	\includegraphics[scale=0.43]{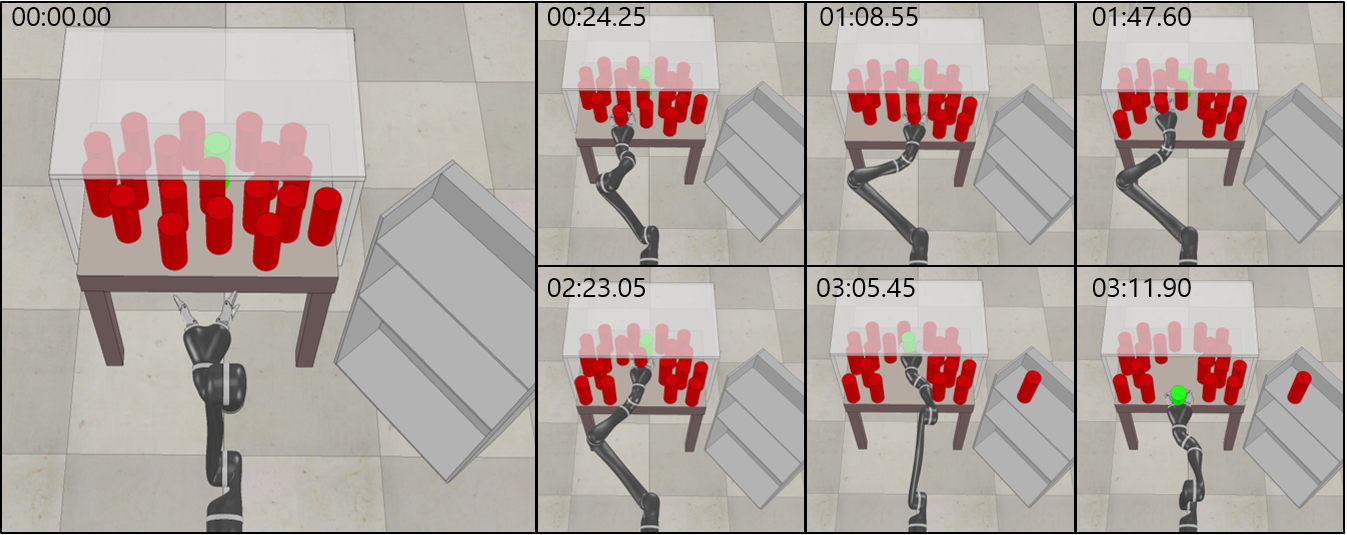}
    \caption{Snapshots of the experiment with 20 objects. Four objects are relocated until the target is grasped. Elapsed time is shown at the left-top.}
    \label{fig:snapshot}
\end{figure*}

The results show that our proposed method takes the shortest time to complete a grasping task in average. Also, the number of removed objects is the minimum among all methods. Specifically, the number of obstacles relocated is improved by $35\%$ compared to the baseline method and improved by $26\%$ compared to the Gaussian method. The execution time is improved by $31\%$ and $19\%$, respectively.
The promising result mainly comes from the way that our method finds the direction to approach to the target, which takes the direction with the lowest density of obstacles. In addition, our method can avoid local minima by changing the distance $d_{\mbox{\scriptsize max}}$ depending on the current configuration of objects. Thus, our method is able to prevent grasping objects that are not necessarily removed to produce a collision-free path. 

Since the object configuration is very dense, the straight path generated by the baseline method passes through an area occupied by many obstacles. Therefore, it turns out that most of the obstacles have to be removed to make a collision-free path. On the other hand, the Gaussian method removes a smaller number of obstacles compared to the baseline. However, it often removes obstacles more than necessary because it could fall into local minima if a threshold value necessary for executing the method is chosen inappropriately. For this reason, the result of the Gaussian method is worse than the baseline method when $N=8$.

In addition to the comparative study, we run experiments in more difficult situations. We attach video clips as supplementary materials showing an experiment with 20 objects (snapshots shown in Fig.~\ref{fig:snapshot}) and another experiment where object locations change dynamically during execution. The ability to work in such difficult settings shows robustness of our method.

\section{Conclusions}

In this paper, we propose a planning algorithm for grasping a target in clutter. The algorithm finds a sequence of obstacles to be relocated to make a collision-free path. Our method aims to minimize the number of obstacles to be relocated for time (or energy) efficient performance. We provide evaluations of our algorithm in a realistic simulated environment. The experiments show that our proposed method outperforms competitive existing methods. Overall, the contributions of our work include the following: (1) The algorithm is shown to be complete and runs in polynomial time. As a result, our method can find the obstacles to be removed in real time (i.e., very short planning time) and guaranteed to grasp a target object which could be in the middle of densely located obstacles. (2) Although the proposed method does not have guarantees for solution quality (like other work), the empirical results compared with other competitive methods show that our method produce high quality solutions in terms of the execution time. (3) Our modification of VFH+ removes threshold values that may be adjusted during execution so reduces manual labor in exploring different values of the thresholds. (4) Our method can deal with dynamically changing object locations and a large number of objects up to 20.

In the future, we plan to develop a method for grasping objects in clutter where the configuration of the objects is partially observable. The manipulator should deal with the uncertainties regarding the poses of the objects. Also, we plan to consider non-prehensile actions like pushing and pulling for the cases where some objects could not be grasped or objects just need to be moved slightly. Another direction to study is considering different grasping poses like picking objects from the top. Lastly, we will consider asymmetric objects so simply approaching to objects from any arbitrary direction would not succeed.



\bibliographystyle{IEEEtran}
\bibliography{references}

\end{document}